\documentclass{article}

\usepackage{tabularx}
\usepackage{needspace}
\usepackage{placeins}
\usepackage{booktabs}
\usepackage{multirow}
\usepackage{enumitem}
\usepackage{algorithm}
\usepackage{algpseudocode}
\usepackage{amsmath}

\usepackage{makecell}
\usepackage{graphicx}
\usepackage{xcolor}
\usepackage[table]{xcolor}

\usepackage{threeparttable}
\usepackage{siunitx}

\usepackage{microtype}
\usepackage{graphicx}
\usepackage{subcaption}
\usepackage{booktabs} 

\usepackage{hyperref}

\makeatletter
\@namedef{ver@algorithmic.sty}{9999/12/31}
\makeatother
\usepackage[accepted]{icml2026}

\usepackage{amsmath}
\usepackage{amssymb}
\usepackage{mathtools}
\usepackage{amsthm}

\usepackage[capitalize,noabbrev]{cleveref}

\theoremstyle{plain}

\theoremstyle{definition}

\theoremstyle{remark}

\usepackage[textsize=tiny]{todonotes}

\icmltitlerunning{Rethinking Small VLM Quantization: From Component-Wise Analysis to Hardware-Aware Edge Deployment}

\begin{document}

\twocolumn[
  \icmltitle{Rethinking Small VLM Quantization: From Component-Wise Analysis to Hardware-Aware Edge Deployment}



  \icmlsetsymbol{equal}{*}

  \begin{icmlauthorlist}
    \icmlauthor{Hyeju Shin}{equal,yyy}
    \icmlauthor{Chorwon Kim}{equal,yyy}
    \icmlauthor{Ryangsoo Kim}{yyy}
    \icmlauthor{Hark Yoo}{yyy}
    \icmlauthor{Jaein Kim}{yyy}
  \end{icmlauthorlist}

  \icmlaffiliation{yyy}{ETRI, Republic of Korea}

  \icmlcorrespondingauthor{Hark Yoo}{harkyoo@etri.re.kr}
  \icmlcorrespondingauthor{Jaein Kim}{jaein@etri.re.kr}

  \icmlkeywords{vision-language models, post-training quantization, component-wise analysis, edge deployment, energy-efficient inference}

  \vskip 0.3in
]



\printAffiliationsAndNotice{}  

\begin{abstract}
    The emergence of vision language models with fewer than 3 billion parameters has accelerated the implementation of on-device multimodal intelligence. However, a detailed understanding of component-wise quantization remains a bottleneck for optimal deployment. This paper presents a systematic evaluation framework for empirically validating five hypotheses across six quantization configurations on the Jetson Orin NX and AGX. By separating the vision encoder, projector, and large language model backbone yields the following results: (1) Quantization sensitivity is governed by the structural paradigm (MoE vs. dense) rather than scale alone, with MoE backbones mitigating INT4 noise where dense backbones degrade; (2) SigLIP encoders incur disproportionate INT8 latency on Jetson Ampere--a deployment-specific encoder-kernel-hardware interaction, not a SigLIP flaw; (3) Although INT4 quantization of LLMs greatly reduces VRAM consumption, it also causes slower token generation due to dequantization overhead; (4) Composite quantization errors are largely additive, except along the modality-alignment path, which is architecture-dependent; (5) The intelligence-per-joule profile varies significantly across platforms owing to memory bandwidth constraints.
\end{abstract}

\begin{table*}[t]
\caption{Comprehensive evaluation: perception, cognition, and total MME scores with percentage variations relative to \texttt{cfg0} (baseline) across all models and configurations.}
\label{tab:final-complete-table}
\centering
\begin{threeparttable}
\resizebox{\textwidth}{!}{
\begin{tabular}{ll ccc c ccc}
\toprule
\multirow{2}{*}{\textbf{Model}} & \multirow{2}{*}{\textbf{Config}} & \multicolumn{3}{c}{\textbf{Jetson NX Orin}} & & \multicolumn{3}{c}{\textbf{Jetson AGX Orin}} \\
\cmidrule{3-5} \cmidrule{7-9}
& & \textbf{Perc. ($\Delta\%$)} & \textbf{Cogn. ($\Delta\%$)} & \textbf{Total ($\Delta\%$)} & & \textbf{Perc. ($\Delta\%$)} & \textbf{Cogn. ($\Delta\%$)} & \textbf{Total ($\Delta\%$)} \\
\midrule
\multirow{6}{*}{Qwen3-VL-2B-Instruct}
& cfg0 (Base) & 1508.20 & 513.57 & 2021.78 & & 1485.46 & 497.86 & 1983.32 \\
& cfg1 & 1524.25 (\textcolor{blue}{+1.06\%}) & 553.57 (\textcolor{blue}{+7.79\%}) & 2077.82 (\textcolor{blue}{+2.77\%}) & & 1535.48 (\textcolor{blue}{+3.37\%}) & 533.57 (\textcolor{blue}{+7.17\%}) & 2069.05 (\textcolor{blue}{+4.32\%}) \\
& cfg2 & 1512.75 (\textcolor{blue}{+0.30\%}) & 477.14 (\textcolor{red}{-7.09\%}) & 1989.89 (\textcolor{red}{-1.58\%}) & & 1501.77 (\textcolor{blue}{+1.10\%}) & 503.57 (\textcolor{blue}{+1.15\%}) & 2005.34 (\textcolor{blue}{+1.11\%}) \\
& cfg3 & 1514.92 (\textcolor{blue}{+0.45\%}) & 531.43 (\textcolor{blue}{+3.48\%}) & 2046.35 (\textcolor{blue}{+1.22\%}) & & 1538.39 (\textcolor{blue}{+3.56\%}) & 526.07 (\textcolor{blue}{+5.67\%}) & 2064.46 (\textcolor{blue}{+4.09\%}) \\
& cfg4 & 1505.65 (\textcolor{red}{-0.17\%}) & 467.86 (\textcolor{red}{-8.90\%}) & 1973.51 (\textcolor{red}{-2.39\%}) & & 1504.48 (\textcolor{blue}{+1.28\%}) & 523.93 (\textcolor{blue}{+5.24\%}) & 2028.41 (\textcolor{blue}{+2.27\%}) \\
& cfg5 & 1511.96 (\textcolor{blue}{+0.25\%}) & 517.14 (\textcolor{blue}{+0.69\%}) & 2029.10 (\textcolor{blue}{+0.36\%}) & & 1524.09 (\textcolor{blue}{+2.60\%}) & 554.64 (\textcolor{blue}{+11.40\%}) & 2078.73 (\textcolor{blue}{+4.81\%}) \\
\midrule
\multirow{6}{*}{DeepSeek-VL2-Tiny}
& cfg0 (Base) & 1550.76 & 360.36 & 1911.12 & & 1556.35 & 360.36 & 1916.71 \\
& cfg1 & 1584.33 (\textcolor{blue}{+2.16\%}) & 368.21 (\textcolor{blue}{+2.18\%}) & 1952.54 (\textcolor{blue}{+2.17\%}) & & 1589.25 (\textcolor{blue}{+2.11\%}) & 372.86 (\textcolor{blue}{+3.47\%}) & 1962.10 (\textcolor{blue}{+2.37\%}) \\
& cfg2 & 1552.34 (\textcolor{blue}{+0.10\%}) & 364.64 (\textcolor{blue}{+1.19\%}) & 1916.98 (\textcolor{blue}{+0.31\%}) & & 1548.53 (\textcolor{red}{-0.50\%}) & 376.07 (\textcolor{blue}{+4.36\%}) & 1924.60 (\textcolor{blue}{+0.41\%}) \\
& cfg3 & 1582.56 (\textcolor{blue}{+2.05\%}) & 370.36 (\textcolor{blue}{+2.78\%}) & 1952.92 (\textcolor{blue}{+2.19\%}) & & 1580.41 (\textcolor{blue}{+1.55\%}) & 370.36 (\textcolor{blue}{+2.78\%}) & 1950.76 (\textcolor{blue}{+1.78\%}) \\
& cfg4 & 1545.90 (\textcolor{red}{-0.31\%}) & 351.07 (\textcolor{red}{-2.58\%}) & 1896.98 (\textcolor{red}{-0.74\%}) & & 1546.17 (\textcolor{red}{-0.65\%}) & 363.57 (\textcolor{blue}{+0.89\%}) & 1909.74 (\textcolor{red}{-0.36\%}) \\
& cfg5 & 1574.16 (\textcolor{blue}{+1.51\%}) & 370.36 (\textcolor{blue}{+2.78\%}) & 1944.51 (\textcolor{blue}{+1.75\%}) & & 1584.90 (\textcolor{blue}{+1.83\%}) & 368.21 (\textcolor{blue}{+2.18\%}) & 1953.11 (\textcolor{blue}{+1.90\%}) \\
\midrule
\multirow{6}{*}{PaliGemma2-3B}
& cfg0 (Base) & 1424.94 & 236.07 & 1661.01 & & 1429.53 & 228.57 & 1658.10 \\
& cfg1 & 1387.46 (\textcolor{red}{-2.63\%}) & 211.43 (\textcolor{red}{-10.44\%}) & 1598.89 (\textcolor{red}{-3.74\%}) & & 1389.45 (\textcolor{red}{-2.80\%}) & 209.29 (\textcolor{red}{-8.44\%}) & 1598.74 (\textcolor{red}{-3.58\%}) \\
& cfg2 & 1425.10 (\textcolor{blue}{+0.01\%}) & 238.57 (\textcolor{blue}{+1.06\%}) & 1663.68 (\textcolor{blue}{+0.16\%}) & & 1424.90 (\textcolor{red}{-0.32\%}) & 228.57 (+0.00\%) & 1653.47 (\textcolor{red}{-0.28\%}) \\
& cfg3 & 1390.95 (\textcolor{red}{-2.39\%}) & 209.29 (\textcolor{red}{-11.34\%}) & 1600.24 (\textcolor{red}{-3.66\%}) & & 1391.08 (\textcolor{red}{-2.69\%}) & 209.29 (\textcolor{red}{-8.44\%}) & 1600.37 (\textcolor{red}{-3.48\%}) \\
& cfg4 & 1429.35 (\textcolor{blue}{+0.31\%}) & 238.93 (\textcolor{blue}{+1.21\%}) & 1668.28 (\textcolor{blue}{+0.44\%}) & & 1420.90 (\textcolor{red}{-0.60\%}) & 236.07 (\textcolor{blue}{+3.28\%}) & 1656.97 (\textcolor{red}{-0.07\%}) \\
& cfg5 & 1382.68 (\textcolor{red}{-2.97\%}) & 207.14 (\textcolor{red}{-12.25\%}) & 1589.83 (\textcolor{red}{-4.29\%}) & & 1379.90 (\textcolor{red}{-3.47\%}) & 209.29 (\textcolor{red}{-8.44\%}) & 1589.19 (\textcolor{red}{-4.16\%}) \\
\midrule
\multirow{6}{*}{LLaVA-OV-0.5B}
& cfg0 (Base) & 1210.76 & 151.07 & 1361.83 & & 1214.02 & 155.71 & 1369.73 \\
& cfg1 & 967.52 (\textcolor{red}{-20.09\%}) & 174.29 (\textcolor{blue}{+15.37\%}) & 1141.81 (\textcolor{red}{-16.16\%}) & & 1029.01 (\textcolor{red}{-15.24\%}) & 184.64 (\textcolor{blue}{+18.58\%}) & 1213.65 (\textcolor{red}{-11.39\%}) \\
& cfg2 & 1209.34 (\textcolor{red}{-0.12\%}) & 151.07 (+0.00\%) & 1360.41 (\textcolor{red}{-0.1\%}) & & 1215.06 (\textcolor{blue}{+0.09\%}) & 148.21 (\textcolor{red}{-4.82\%}) & 1363.28 (\textcolor{red}{-0.47\%}) \\
& cfg3 & 965.8 (\textcolor{red}{-20.23\%}) & 171.79 (\textcolor{blue}{+13.72\%}) & 1137.59 (\textcolor{red}{-16.47\%}) & & 1021.05 (\textcolor{red}{-15.90\%}) & 186.43 (\textcolor{blue}{+19.73\%}) & 1207.48 (\textcolor{red}{-11.85\%}) \\
& cfg4 & 1220.27 (\textcolor{blue}{+0.79\%}) & 148.93 (\textcolor{red}{-1.42\%}) & 1369.20 (\textcolor{blue}{+0.54\%}) & & 1212.08 (\textcolor{red}{-0.16\%}) & 153.21 (\textcolor{red}{-1.61\%}) & 1365.30 (\textcolor{red}{-0.32\%}) \\
& cfg5 & 974.79 (\textcolor{red}{-19.49\%}) & 174.29 (\textcolor{blue}{+15.37\%}) & 1149.08 (\textcolor{red}{-15.62\%}) & & 1026.16 (\textcolor{red}{-15.47\%}) & 189.29 (\textcolor{blue}{+21.57\%}) & 1215.45 (\textcolor{red}{-11.26\%}) \\
\midrule
\multirow{6}{*}{Kosmos-2.5}
& cfg0 (Base) & 464.80 & 197.50 & 662.30 & & 464.55 & 197.50 & 662.05 \\
& cfg1 & 454.97 (\textcolor{red}{-2.11\%}) & 196.79 (\textcolor{red}{-0.36\%}) & 651.76 (\textcolor{red}{-1.59\%}) & & 453.05 (\textcolor{red}{-2.48\%}) & 196.79 (\textcolor{red}{-0.36\%}) & 649.84 (\textcolor{red}{-1.84\%}) \\
& cfg2 & 466.05 (\textcolor{blue}{+0.27\%}) & 198.21 (\textcolor{blue}{+0.36\%}) & 664.27 (\textcolor{blue}{+0.30\%}) & & 465.80 (\textcolor{blue}{+0.27\%}) & 197.50 (+0.00\%) & 663.30 (\textcolor{blue}{+0.19\%}) \\
& cfg3 & 457.14 (\textcolor{red}{-1.65\%}) & 196.79 (\textcolor{red}{-0.36\%}) & 653.92 (\textcolor{red}{-1.27\%}) & & 459.97 (\textcolor{red}{-0.99\%}) & 196.79 (\textcolor{red}{-0.36\%}) & 656.76 (\textcolor{red}{-0.80\%}) \\
& cfg4 & 465.05 (\textcolor{blue}{+0.05\%}) & 201.43 (\textcolor{blue}{+1.99\%}) & 666.48 (\textcolor{blue}{+0.63\%}) & & 464.05 (\textcolor{red}{-0.11\%}) & 198.21 (\textcolor{blue}{+0.36\%}) & 663.27 (\textcolor{blue}{+0.18\%}) \\
& cfg5 & 457.39 (\textcolor{red}{-1.59\%}) & 199.29 (\textcolor{blue}{+0.91\%}) & 656.67 (\textcolor{red}{-0.85\%}) & & 453.80 (\textcolor{red}{-2.31\%}) & 196.79 (\textcolor{red}{-0.36\%}) & 650.59 (\textcolor{red}{-1.73\%}) \\
\bottomrule
\end{tabular}
}
\begin{tablenotes}[flushleft] 
    \scriptsize
    \item[] \textcolor{black}{\textbf{Note:} Each data point reflects the average of $n \ge 3$ independent runs. Refer to Appendix~\ref{app:mme-score-all-appendix} for the detailed bar chart representing the mean score across runs.}
\end{tablenotes}
\end{threeparttable}
\end{table*}

\begin{table}[t]
\caption{Definition of configurations for component-wise quantization sensitivity analysis.}
\label{tab:hypotheses-optimized}
\begin{center}
\small
\begin{tabularx}{\columnwidth}{@{} l >{\centering\arraybackslash}p{0.9cm} >{\centering\arraybackslash}p{1.0cm} >{\centering\arraybackslash}p{0.9cm} X @{}}
\toprule
\textbf{Config} & \textbf{Vision} & \textbf{Projector} & \textbf{LLM} & \textbf{Description} \\
\midrule
cfg0 (Base) & FP16 & FP16 & FP16 & Baseline \\
\addlinespace[0.3em]
cfg1 & FP16 & FP16 & \textbf{INT4} & LLM Robustness \\
\addlinespace[0.3em]
cfg2 & FP16 & \textbf{INT8} & FP16 & Proj. Sensitivity \\
\addlinespace[0.3em]
cfg3 & FP16 & \textbf{INT8} & \textbf{INT4} & cfg1 + cfg2 \\
\addlinespace[0.3em]
cfg4 & \textbf{INT8} & FP16 & FP16 & Enc. Sensitivity \\
\addlinespace[0.3em]
cfg5 & \textbf{INT8} & FP16 & \textbf{INT4} & cfg1 + cfg4 \\
\bottomrule
\end{tabularx}
\end{center}
\vskip -0.15in
\end{table}

\section{Introduction}
Recently, large multimodal models (LMMs) have opened a new horizon in artificial intelligence (AI) research owing to their innovative reasoning capabilities, which combine visual understanding and language generation \cite{liu2023visual, sun2024generative, hurst2024gpt}. Small vision-language models (sVLMs) with 3 billion or less parameters are rapidly emerging as practical candidates for on-device AI. These models can be used in mobile, offline, privacy-protected, and low-power environments \cite{chu2023mobilevlm,zhou2024tinyllava,lin2026moe}. However, heterogeneous edge system-on-chip (SoC) environments such as the NVIDIA Jetson Orin series are subject to fundamental hardware constraints, including strict memory bandwidth (NX: 102.4 GB/s; AGX: 204.8 GB/s) and limited VRAM capacity (NX: 16 GB; AGX: 64 GB). In such an environment, model quantization is gaining attention as an essential compression technique to ensure practical reasoning capacity \cite{kim2022integer,cai2023efficientvit,xu2024survey}.

However, most existing research on the quantization of large language models (LLMs) and vision-language models (VLMs) has generally focused on evaluating the entire model end-to-end in high-performance server-grade GPU environments \cite{lee2024exploring,xiao2023smoothquant,frantar2022gptq,dettmers2022gpt3}. Existing studies have yet to provide a systematic basis for key deployment-time decisions faced by engineers, including which components should be prioritized for quantization and how bit-widths should be allocated across these components. A VLM typically consists of three major components: a vision encoder that extracts visual features, a projector that maps these features into the embedding space of the language model, and an LLM backbone that performs the final inference \cite{liu2023visual,alayrac2022flamingo,chen2024sharegpt4v}. Each component has different parameter sizes, model structures, and computational characteristics; thus, the sensitivity of each component to quantization should be evaluated independently. Studies that control and quantify the marginal contribution of each component are rare \cite{sharshar2025vision}. Cross-platform comparative studies, especially on heterogeneous edge SoCs, are even rarer.

Motivated by our preliminary observations, we formulate five hypotheses and evaluate each by designing a hypothesis-driven evaluation framework. We evaluate five architecturally diverse sVLMs: Qwen3-VL-2B, DeepSeek-VL2-Tiny, PaliGemma2-3B, LLaVA-OV-0.5B, and Kosmos-2.5. The developed configurations include single-component ablations (\texttt{cfg1}, \texttt{cfg2}, and \texttt{cfg4}), which isolate the effect of quantizing individual components, and composite settings (\texttt{cfg3} and \texttt{cfg5}), which capture interactions among jointly quantized components. We conduct all experiments on the Jetson Orin NX and Jetson AGX platforms and analyze each configuration along four axes: MME benchmark accuracy, VRAM footprint, component-wise latency across the vision encoder, projector, and LLM time per output token (TPOT), and intelligence-per-joule (IPJ)\cite{saad2025intelligence}.

We formulate the following hypotheses:
\begin{itemize}
  \item \textbf{Scale-Dependent Sensitivity according to Parameter Size (H1):} INT4 quantization of the LLM backbone results in increasingly severe accuracy degradation as the model size decreases. This effect is expected to be especially pronounced in ultra-small sVLMs with fewer than 1B parameters, where limited representational redundancy makes the model less robust to quantization noise.
  \item \textbf{Vision Encoder Architecture-Specific Latency Anomaly (H2):} SigLIP-based vision encoder quantization disproportionately increases the latency relative to the variation in accuracy because of interactions between certain architectures and the quantization software stack that are specific to the hardware.
  \item \textbf{Relationship between Resource Savings and Inference Efficiency (H3):} BitsAndBytes-based LLM INT4 quantization achieves over 50\% VRAM savings, whereas TPOT consistently increases due to the dequantization overhead.
  \item \textbf{Non-Additive Interaction of Composite Quantization Errors (H4):} The performance degradation when quantizing heterogeneous modalities is characterized by a non-additive pattern that cannot be explained by simple arithmetic summation of the individual component errors.
  \item \textbf{Stability and Transferability of Performance Rankings across Platforms (H5):} The accuracy ranking of the models is platform-invariant across the quantization configurations and heterogeneous edge platforms (NX vs. AGX), whereas the latency and energy profiles are platform-specific.
\end{itemize}

The main contributions of this study are as follows: \textbf{(1)} We present the first systematic controlled experiment that independently separates the vision encoder, projector, and LLM backbone of VLMs using 3 billion parameters to measure accuracy sensitivity and resource efficiency. \textbf{(2)} Through cross-hardware evaluation using two heterogeneous platforms (Jetson Orin NX and AGX), we empirically distinguish platform-consistent accuracy rankings from hardware-specific latency profiles. \textbf{(3)} We identify abnormal latency phenomena that occur when applying vision INT8 quantization in SigLIP-based architectures, and analyze these issues from a hardware-software stack perspective. \textbf{(4)} By testing the hypotheses, we provide practical deployment guidelines for implementing modality- and hardware-aware precision allocation strategies in edge environments.

\section{Related Works}
\textbf{Small Vision-Language Models.} A sVLM is a multimodal AI system with 7 billion parameters or less. sVLMs are designed to efficiently process and integrate visual and textual information while supporting implementation in resource-constrained environments, such as edge and mobile devices \cite{chu2023mobilevlm}. A typical VLM architecture consists of three core components: a vision encoder that extracts visual features from images, a projector that maps visual representations into the embedding space of the language model, and a LLM backbone for text generation and reasoning \cite{liu2023visual}. These components serve distinct functions within the model and may, therefore, differ in their sensitivity to quantization \cite{sharshar2025vision}. Table~\ref{tab:refined-model-details} summarizes the architectural configurations of the five models selected for this study. Except for Kosmos-2.5, the other four models use SigLIP as the vision encoder. Thus, SigLIP can be treated as a controlled architectural factor in the architecture-effect analysis for RQ2, as demonstrated in Section 4.2.

\textbf{Quantization of VLMs.} Model quantization is a widely used compression technique that reduces the memory footprint and computational overhead by representing model weights and activations in low-precision formats. Existing quantization methods, such as activation-aware weight quantization (AWQ), generative pre-trained transformer quantization (GPTQ), SmoothQuant, and BitsAndBytes-based INT8/INT4 quantization, have demonstrated substantial efficiency gains in large-scale neural models \cite{lin2024awq,frantar2022gptq,xiao2023smoothquant,dettmers2022gpt3}.

These techniques have been extensively explored for LLMs. Nevertheless, their extension to VLMs is less straightforward. VLMs combine multiple heterogeneous components, including a vision encoder, a cross-modal projector, and an LLM backbone. Because these components differ in size, function, and sensitivity to numerical precision, uniform quantization may not provide the best trade-off between efficiency and accuracy cost \cite{sharshar2025vision,xue2025vlmq,li2025mbq}. This motivates component-wise mixed-precision quantization, where different bit-widths are assigned to different VLM components based on their quantization sensitivity and deployment requirements.

\section{Experimental Setup}
\subsection{Hardware Setup}
\label{ssec:ex_hw_setup}
The Jetson Orin family is an edge AI platform built on NVIDIA Ampere architecture. It offers multiple memory configurations, such as the NX (16 GB, LPDDR5) and AGX (64 GB, LPDDR5). Even with the same quantization configuration, the latency profile can vary across hardware platforms depending on the level of optimization of the kernels. INT4 quantization on certain platforms, for instance, involves dequantization overhead when using BitsAndBytes. If native INT4 operation support is limited in the Ampere architecture of Jetson Orin, the TPOT may increase \cite{xu2024survey}. This explains why VRAM savings do not necessarily correspond to latency improvements, as confirmed by the data presented in Section 4.3.

\begin{table}[t]
\caption{Hardware specifications of Jetson Orin platforms}
\label{tab:jetson-specs-final}
\begin{center}
\small
\begin{tabularx}{\columnwidth}{@{} l X X @{}}
\toprule
\textbf{Specification} & \textbf{Jetson Orin NX} & \textbf{Jetson Orin AGX} \\
\midrule
\textbf{GPU} & 1024-core Ampere, 32 Tensor Cores & 2048-core Ampere, 64 Tensor Cores \\
\addlinespace[0.2em]
\textbf{CPU} & 8-core ARM CPU & 12-core ARM CPU \\
\addlinespace[0.2em]
\textbf{Memory} & 16GB LPDDR5 & 64GB LPDDR5 \\
\addlinespace[0.2em]
\textbf{Memory BW} & 102.4 GB/s & 204.8 GB/s \\
\addlinespace[0.2em]
\textbf{TDP} & 25 W (MODE\_25W) & 50 W (MODE\_50W) \\
\addlinespace[0.2em]
\textbf{AI Perf.} & 157 TOPS & 275 TOPS \\
\addlinespace[0.2em]
\textbf{Jetpack} & 6.2.1 & 6.2.1 \\
\bottomrule
\end{tabularx}
\end{center}
\vskip -0.3in
\end{table}

The Jetson Orin family comprises the Jetson Orin Nano, NX, and AGX. The Jetson Orin Nano was excluded from the experimental set because it consistently encountered out-of-memory (OOM) failures during VLM inference. Prior work has shown that these failures stem from the substantial peak memory overhead introduced during the quantized weight conversion stage in BitsAndBytes \cite{dettmers2022gpt3}. This emphasizes the need to provide additional memory capacity and bandwidth at runtime, beyond that required to simply load the model parameters. Consequently, experiments were carried out only on the Jetson Orin NX and AGX platforms, as summarized in Table~\ref{tab:jetson-specs-final} \cite{nvidia_jetson_orin}.

\subsection{Models}
\label{ssec:ex_models}
This study focuses on five 3B-parameter models supported by the VLMEvalKit framework, which is also employed in the Open VLM Leaderboard \cite{open_vlm_leaderboard}. These models were chosen based on their frequency of academic citation and prevalence in industrial applications. The main selection criteria are architectural diversity (SigLIP vs. non-SigLIP) \cite{zhai2023sigmoid} and variability in the LLM backbones, as shown in Table~\ref{tab:refined-model-details} \cite{bai2025qwen3,wu2024deepseek,steiner2024paligemma,li2024llava,lv2024kosmos25}.

\begin{table*}[t]
\caption{Detailed architectural specifications and parameter distributions across five multimodal models. All percentages represent the relative share of the total model parameters. For Kosmos-2.5, the output projection is tied to the input embedding, so no separate lm\_head parameters are counted.} 
\label{tab:refined-model-details}
\vskip -0.05in
\begin{center}
\footnotesize 
\begin{tabularx}{\textwidth}{@{} l l X l X l X @{}}
\toprule
\multirow{2.5}{*}{\textbf{Model}} & \multicolumn{2}{c}{\textbf{Vision Encoder}} & \multicolumn{2}{c}{\textbf{Language Model}} & \multicolumn{2}{c}{\textbf{Projector}} \\
\cmidrule(lr){2-3} \cmidrule(lr){4-5} \cmidrule(lr){6-7}
& \textbf{Arch} & \textbf{Params Details} & \textbf{Arch} & \textbf{Params Details} & \textbf{Arch} & \textbf{Params Details} \\
\midrule

Qwen3-VL-2B & \multirow{2}{*}{SigLIP-2} & $\sim$306M (12.6\%) & Qwen3 & $\sim$1.72B (70.6\%) & MLP-based & $\sim$100M (4.13\%) \\
-Instruct (2.44B) & & \tiny Linear: 302M Conv3d: 1.6M & (Dense/MoE) & \tiny lm\_head: 311M (12.8\%) & Merger & \tiny \\
\addlinespace[0.5em]

DeepSeek-VL2 & SigLIP- & $\sim$428M (12.7\%) & DeepSeek & $\sim$2.93B (87.1\%) & 2-layer & $\sim$7.53M (0.22\%) \\
-Tiny (3.37B) & So400m & \tiny Linear: 426M ViT: 0.8M & -MoE & \tiny MoE-Gate: 0.9M & MLP & \tiny \\
\addlinespace[0.5em]

PaliGemma2-3B & SigLIP- & $\sim$412M (11.4\%) & \multirow{2}{*}{Gemma 2} & $\sim$2.61B (72.2\%) & Linear & $\sim$2.65M (0.07\%) \\
(3.6B) & So400m & \tiny Linear: 411M Conv2d: 0.7M & & \tiny lm\_head: 592M (16.4\%) & Proj. & \tiny \\
\addlinespace[0.5em]

LLaVA-OV-0.5B & SigLIP- & $\sim$397M (38.6\%) & \multirow{2}{*}{Qwen-2} & $\sim$494M (48.0\%) & 2-layer & $\sim$1.83M (0.18\%) \\
(1.03B) & So400m & \tiny Linear: 396M Conv2d: 0.7M & & \tiny lm\_head: 136M (13.2\%) & MLP & \tiny \\
\addlinespace[0.5em]

Kosmos-2.5 & Pix2Struct & $\sim$512.8M (37.31\%) & Decoder-only & $\sim$846.9M (61.61\%) & \multirow{2}{*}{Resampler} & $\sim$14.9M (1.09\%) \\
(1.37B) & -based ViT & \tiny Linear: 512M & Trans. & \tiny lm\_head: tied & & \tiny \\
\bottomrule
\end{tabularx}
\end{center}
\vskip -0.2in
\end{table*}

\begin{algorithm}[htb!]
    \caption{Component Latency \& VRAM Profiling for sVLM}
    \label{alg:profiling}
    \begin{algorithmic}[1]
        \renewcommand{\algorithmicrequire}{\textbf{Input:}}
        \renewcommand{\algorithmicensure}{\textbf{Output:}}
        
        \Require Quantized VLM $\mathcal{M}_Q = \{V, P, L\}$, Evaluation Dataset $D$, Output token length $N$
        \Ensure Accuracy Score, Peak VRAM, Latency components ($L_V, L_P, L_{TPOT}$)

        \Statex
        \Statex \textit{\# 1. Accuracy and Peak VRAM Measurement}
        \State ResetCUDAMemoryStats()\;
        \State $Score \leftarrow \text{VLMEvalKit\_Evaluate}(\mathcal{M}_Q, D)$\;
        \State $Peak\_VRAM \leftarrow \text{GetMaxMemoryAllocated()}$\;

        \Statex 
        \Statex \textit{\# 2. Latency Analysis}
        \State $I, X \leftarrow \text{Sample}(D)$ \Comment{Image $I$, Text Prompt $X$}
        \Statex
        \State CUDASynchronize()\;
        \State $t_0 \leftarrow \text{GetCurrentTime()}$\;
        \State $E_v \leftarrow \mathcal{M}_Q.V(I)$\;
        \State CUDASynchronize()\;
        \State $t_1 \leftarrow \text{GetCurrentTime()}$\;
        \State $L_V \leftarrow t_1 - t_0$ \Comment{Vision Encoder Latency}

        \Statex 
        \State $E_p \leftarrow \mathcal{M}_Q.P(E_v)$\;
        \State CUDASynchronize()\;
        \State $t_2 \leftarrow \text{GetCurrentTime()}$\;
        \State $L_P \leftarrow t_2 - t_1$ \Comment{Projector Latency}

        \Statex 
        \State $Inputs \leftarrow \text{Concat}(E_p, \text{Embed}(X))$\;
        \State $\mathcal{M}_Q.L.\text{Generate}(Inputs, \text{max\_new\_tokens}=1)$\;
        \State CUDASynchronize()\;
        \State $t_3 \leftarrow \text{GetCurrentTime()}$\;

        \Statex 
        \State $Out_N \leftarrow \mathcal{M}_Q.L.\text{G}(Inputs, \text{max\_tokens}=N)$\;
        \State CUDASynchronize()\;
        \State $t_4 \leftarrow \text{GetCurrentTime()}$\;

        \Statex 
        \State $Total\_Decode\_Time \leftarrow t_4 - t_3$\;
        \State $L_{TPOT} \leftarrow Total\_Decode\_Time / (N-1)$ \Comment{TPOT}
        
        \Statex 
    \Return{$Score, Peak\_VRAM, L_V, L_P, L_{TPOT}$}
    \end{algorithmic}
\end{algorithm}

\subsection{Datasets}
\label{ssec:ex_data}
For evaluation, we use the multimodal large language model evaluation (MME) benchmark, which is the most widely adopted benchmark for VLM models. The MME scores each subtask by combining the ACC (accuracy per question) and ACC+ (granted when both questions for a single image are answered correctly). The overall maximum score is 2,800 points, calculated as the sum of the cognition and perception scores \cite{fu2023mme}.

\subsection{Quantization Methods}
\label{ssec:ex_qntz_mth}
In this study, we applied 4-bit (INT4) quantization exclusively to the LLM to reduce VRAM consumption, given that the LLM typically represents more than 70\% of the total parameters in most VLMs (Table~\ref{tab:refined-model-details}). Vision-related components (the vision encoder and projector), which are more susceptible to degradation arising from information loss, were kept at INT8 or higher precision. This prevents performance degradation in terms of modality alignment performance. Based on this design, we defined six quantization configurations: three single-component quantization ablations (\texttt{cfg1}, \texttt{cfg2}, and \texttt{cfg4}), in which two components are fixed at FP16 and one component is quantized; and two composite quantizations (\texttt{cfg3} and \texttt{cfg5}), in which both components are quantized simultaneously. This design incorporates structural pairs that enable independent measurement of the limiting effects of each component and verification of the additivity in composite quantization configurations. The detailed experimental configurations are presented in Table~\ref{tab:hypotheses-optimized}.

\subsection{Evaluation Metrics}
\label{ssec:ex_eval_metric}
We utilized the open-source VLMEvalKit framework for accuracy evaluation \cite{duan2024vlmevalkit}. The latency was assessed by decomposing it into three parts: vision encoding, projector, and LLM TPOT. The VRAM consumption was reported as the peak usage (GB). The pseudocode used for reproduction is provided in Algorithm 1.

Algorithm 1 is built to precisely pinpoint performance bottlenecks in edge environments. The peak VRAM indicates increases in temporary memory usage introduced by the BitsAndBytes quantization model during dequantization in the inference phase. It also reflects the actual hardware capacity. In addition, the latency was decomposed and measured around each component (vision encoder, projector, and LLM) by recording the execution times before and after each component. Notably, for the LLM, only TPOT was measured independently, while excluding the latency associated with generating the first token (the prefill phase).

\section{Experimental Results}
In this section, we explore the impact of component-wise quantization of the sVLMs on the performance and hardware profiles in edge SoC environments, such as the Jetson Orin NX/AGX, by examining five research questions (RQs).

\subsection{RQ1: [Scale and Architecture Sensitivity] For different backbone architectures and model scales, how does INT4 quantization of the LLM backbone affect the change in MME accuracy?}
\label{ssec:res_rq1}
\textcolor{black}{The accuracy variations for each model ($\Delta MME = cfg_N - base$) from the single-component ablations (\texttt{cfg1}: LLM INT4; \texttt{cfg2}: Projector INT8; and \texttt{cfg4}: Vision INT8) are summarized in Table~\ref{tab:delta-mme-1col}. The experiments revealed a complex interplay between parameter scale, LLM architecture, and INT4 quantization sensitivity. While the severe degradation in ultra-small models initially appears to align with a scale-dependent trend, a cross-architectural comparison reveals that algorithmic structure dominates over mere parameter scale. Consequently, the empirical evidence rejects the original hypothesis, $\text{H1}$. Reflecting the interaction between architecture and scale, we revise $\text{H1}$ into ($\text{H1}_{\text{rev}}$): The quantization sensitivity of sub-3B VLMs is determined by the underlying structural paradigm (MoE vs. Dense), with parameter scale acting as an aggravating factor only within a homogeneous architecture family.}

\begin{table}[h]
\caption{\textcolor{black}{$\Delta$MME comparison across models under \texttt{cfg1}.}}
\label{tab:delta-mme-1col}
\vskip -0.05in
\centering
\scriptsize 
\renewcommand{\arraystretch}{0.9}
\setlength{\tabcolsep}{0pt}
\begin{tabularx}{\columnwidth}{@{} *{1}{>{\raggedright\arraybackslash}X} *{2}{>{\columncolor{gray!15}\centering\arraybackslash}X} *{2}{>{\centering\arraybackslash}X} @{} }
\toprule
\textbf{Model} & \makecell[c]{\textbf{\# LLM}\\\textbf{Params (B)}} & \textbf{MoE} & \textbf{$\Delta$MME (NX)} & \textbf{$\Delta$MME (AGX)} \\
\midrule
Qwen3-VL & 1.72 & \textbf{O} & +56.04 & +101.88 \\
DeepSeek-VL2 & 2.93 & \textbf{O} & +39.50 & +48.18 \\
PaliGemma2 & 2.61 & \textbf{X} & -62.12 & -59.35 \\
LLaVA-OV & \textbf{0.49} & \textbf{X} & \textbf{-220.02} & \textbf{-156.08} \\
\bottomrule
\end{tabularx}
\end{table}

\needspace{1\baselineskip}
\textbf{Key Findings:}
\begin{itemize}
    \item \textbf{\textcolor{black}{Architecture Dominance Over Scale Sensitivity (Falsification of H1).}} \textcolor{black}{Although the substantial performance decline of LLaVA-OV (Qwen2-0.5B, -220.02 points) initially suggests a scale-dependent vulnerability, this hypothesis is obscured by architectural variables. Notably, the larger model, PaliGemma2 (2.61B), shows a significant decline of 62.12 points, while the smaller MoE-based Qwen3 (1.72B) demonstrates an enhancement in accuracy, gaining 56.04 points. This contradiction rejects \text{H1} based exclusively on scale.}
    \item \textbf{Impact of Architectural Bifurcation on INT4 Accuracy of LLM.} \textcolor{black}{The LLM architecture fundamentally dictates the impact of \texttt{cfg1} on accuracy. Qwen3 (MoE-based, +56.04) and DeepSeek-VL2 (MoE-based, +39.50), achieve consistent improvements across all platforms, whereas the dense PaliGemma2 (Gemma-2B, -62.12) and LLaVA-OV (Qwen2-0.5B, -220.02) suffer pronounced performance declines. This pattern supports a reformulated $\text{H1}_{\text{rev}}$, aligning with prior studies that identify MoE models as notably resilient to severe quantization noise \cite{frantar2023qmoe}}.
\end{itemize}

\subsection{RQ2: [Architectural Latency Anomaly] Does vision INT8 quantization have a significantly different impact on the latency of vision encoding for SigLIP-based and non-SigLIP architectures? Is this latency change independent of the accuracy changes?}
\label{ssec:res_rq2}
The INT8 quantization (\texttt{cfg4}) of the vision encoder resulted in increased latency, which was largely unaffected by changes in the accuracy of the models based on the SigLIP architecture. The latency increase ratio (Table~\ref{tab:vision-latency-optimized}) is calculated as $ratio = cfg4/cfg0$. This ratio was rounded to three decimal places.

\begin{table}[h]
\caption{Impact of vision INT8 quantization on vision encoding latency (ms).}
\label{tab:vision-latency-optimized}
\vskip -0.05in
\centering
\scriptsize
\setlength{\tabcolsep}{0.9pt} 
\begin{tabularx}{\columnwidth}{@{} l l c c >{\columncolor{gray!15}\centering\arraybackslash}p{0.9cm} c c >{\columncolor{gray!15}\centering\arraybackslash}p{0.9cm} c @{}}
\toprule
\multirow{2}{*}{\textbf{Model}} & \multirow{2}{*}{\textbf{\makecell{Vision\\Encoder}}} & \multicolumn{3}{c}{\textbf{Jetson NX}} & \multicolumn{3}{c}{\textbf{Jetson AGX}} & \multirow{2}{*}{\makecell{\textbf{$\Delta$MME}\\\textbf{(NX)}}}\\
\cmidrule(lr){3-5} \cmidrule(lr){6-8}
& & \textbf{cfg0} & \textbf{cfg4} & \textbf{Ratio} $\downarrow$ & \textbf{cfg0} & \textbf{cfg4} & \textbf{Ratio} $\downarrow$ & \\
\midrule
PG2 & \makecell[l]{\textbf{SigLIP-}\\\textbf{So400m}} & 311.9 & 1,203.5 & \textbf{$\times$3.86} & 98.5 & 459.1 & \textbf{$\times$4.66} & +7.3 \\
DS-VL2 & \makecell[l]{\textbf{SigLIP-}\\\textbf{So400m}} & 1,423.3 & 4,297.3 & $\times$3.02 & 903.5 & 2,810.7 & $\times$3.11 & -14.1 \\
LV-OV & \makecell[l]{\textbf{SigLIP-}\\\textbf{So400m}} & 719.8 & 1,879.1 & $\times$2.61 & 381.2 & 925.1 & $\times$2.43 & +7.4 \\
QW3-VL & \textbf{SigLIP-2} & 394.6 & 684.2 & $\times$1.73 & 160.6 & 403.3 & $\times$2.51 & -48.3 \\
Kosmos-2.5 & \makecell[l]{\textbf{Pix2}\\\textbf{Struct}} & 11,029.2 & 12,644.2 & \underline{$\times$1.15} & 2,903.6 & 3,479.4 & \underline{$\times$1.20} & +4.2 \\
\bottomrule
\multicolumn{9}{l}{\textit{Note: Latency is measured in milliseconds (ms).}}
\end{tabularx}
\end{table}

\begin{table}[h]
\caption{Impact of LLM INT4 quantization (\texttt{cfg1}) on VRAM usage and token generation speed (TPOT) relative to \texttt{cfg0} (baseline).}
\label{tab:vram-tpot-final-optimized}
\vskip -0.05in
\centering 
\scriptsize
\setlength{\tabcolsep}{2pt}
\begin{tabularx}{\columnwidth}{@{} l l *{2}{>{\centering\arraybackslash}X} >{\columncolor{gray!15}\centering\arraybackslash}X *{2}{>{\centering\arraybackslash}X} >{\columncolor{gray!15}\centering\arraybackslash}X @{}}
\toprule
\multirow{2}{*}{\textbf{}} & \multirow{2}{*}{\textbf{Models}} & \multicolumn{3}{c}{\textbf{VRAM (GB)} $\downarrow$} & \multicolumn{3}{c}{\textbf{TPOT (ms) $\downarrow$}} \\
\cmidrule(lr){3-5} \cmidrule(lr){6-8}
& & \textbf{cfg0} & \textbf{cfg1} & \textbf{Reduct.} & \textbf{cfg0} & \textbf{cfg1} & \textbf{Incr.} \\
\midrule
\multirow{4}{*}{NX} 
& Qwen3-VL & 4.06 & 2.13 & \textbf{-47.5\%} & 111.1 & 173.1 & \underline{+55.8\%} \\
& PaliGemma2 & 5.85 & 3.13 & -46.5\% & 166.0 & 183.3 & +10.4\% \\
& LLaVA-OV & 1.89 & 1.43 & -24.3\% & 96.5 & 143.9 & +49.1\% \\
\midrule
\multirow{4}{*}{AGX} 
& Qwen3-VL & 4.06 & 2.13 & \textbf{-47.5\%} & 136.2 & 212.4 & +55.9\% \\
& PaliGemma2 & 5.85 & 3.18 & -45.6\% & 137.9 & 182.1 & +32.0\% \\
& LLaVA-OV & 2.25 & 1.76 & -21.8\% & 90.5 & 141.4 & \underline{+56.3\%} \\
\bottomrule
\end{tabularx}
\vskip -0.15in
\end{table}

\needspace{4\baselineskip}
\textbf{Key Findings:}
\begin{itemize}
    \item \textbf{Sharp Increase in Vision INT8 Latency of SigLIP-So400m-Based Models.} When \texttt{cfg4} is applied, the latency of the vision encoders of the three SigLIP-So400m-based models (PaliGemma2, DeepSeek-VL2, and LLaVA-OV) increases sharply (by a factor of 2.43-4.66), compared with that (1.73-2.51-fold) for the SigLIP-2-based Qwen3-VL. The Kosmos-2.5 model with a non-SigLIP architecture exhibits a smaller increase (1.15-1.20-fold). These results support the latency anomaly phenomenon specific to the SigLIP-BitsAndBytes-Ampere interaction established in H2.
    \item \textbf{Independence of Accuracy Changes.} This latency increase occurs regardless of changes in accuracy. For PaliGemma2 and LLaVA-OV, \texttt{cfg4} caused $\Delta$ MME to increase by 7.3 and 7.4, respectively, indicating a slight improvement in accuracy. However, the vision encoding latency increased by factors of 3.86 and 2.61, respectively. Because there was no loss in accuracy, the intuitive judgment that vision INT8 was efficient overlooks the latency cost. This error was confirmed through actual measurements.
    \item \textbf{Hardware-Software Stack Incompatibility.} This incompatibility arises when the BitsAndBytes INT8 kernel fails to efficiently accelerate the computational patterns of the SigLIP vision transformer on the Jetson Orin Ampere architecture \cite{cai2023efficientvit}. Although the Pix2Struct-style Kosmos-2.5 encoder incurs the highest absolute vision latency due to its variable-resolution patching, the relative INT8 overhead remains small precisely because its non-SigLIP computation pattern avoids the kernel fragmentation that BitsAndBytes induces on SigLIP encoders. We verify this with a non-SigLIP control under the same isolated vision-path protocol in Appendix~\ref{app:vision-int8-control}.
\end{itemize}

\subsection{RQ3: [Resource and Accuracy Trade-offs] When applying LLM INT4 quantization, does the VRAM reduction translate into an actual improvement in the token generation latency (TPOT), or is it offset by the dequantization overhead?}
\label{ssec:res_rq3}
Although INT4 quantization of LLMs significantly lowers the VRAM usage, it also reduces the token generation speed due to the extra computational overhead for dequantization. The formulas used to calculate the percentage change of VRAM and TPOT \textcolor{black}{(Table~\ref{tab:vram-tpot-final-optimized})} are as follows:
$$\Delta X = \frac{X_{\text{cfg1}} - X_{\text{cfg0}}}{X_{\text{cfg0}}} \times 100 (\%), \quad X \in \{VRAM, TPOT\}$$

\begin{table}[h]
\caption{\textcolor{black}{Energy consumption (joules) and rate of increase owing to \texttt{cfg1} quantization across Jetson platforms.}}
\label{tab:energy-consumption}
\vskip -0.05in
\centering
\scriptsize 
\setlength{\tabcolsep}{2pt} 
\begin{tabularx}{\columnwidth}{@{} l *{2}{>{\centering\arraybackslash}X} >{\columncolor{gray!15}\centering\arraybackslash}X *{2}{>{\centering\arraybackslash}X} >{\columncolor{gray!15}\centering\arraybackslash}X @{}}
\toprule
\multirow{2}{*}{\textbf{Models}} & \multicolumn{3}{c}{\textbf{NX}} & \multicolumn{3}{c}{\textbf{AGX}} \\
\cmidrule(lr){2-4} \cmidrule(lr){5-7}
& \textbf{cfg0} & \textbf{cfg1} & \textbf{Incr.} $\downarrow$ & \textbf{cfg0} & \textbf{cfg1} & \textbf{Incr.} $\downarrow$ \\
\midrule
Qwen3-VL & 12.31 & 19.04 & \textbf{+54.7\%} & 4.90 & 7.12 & \textbf{+45.3\%} \\
PaliGemma2 & 35.98 & 45.95 & +27.7\% & 13.31 & 16.13 & +21.2\% \\
LLaVA-OV & 21.47 & 21.87 & \underline{+1.9\%} & 21.55 & 21.86 & \underline{+1.4\%} \\
\bottomrule
\multicolumn{7}{l}{\textit{Note: Energy consumption is measured in Joules (J).}}
\end{tabularx}
\end{table}

\textbf{Key Findings:}
\begin{itemize}
    \item \textbf{Trade-off between VRAM Reduction and TPOT Increase.} LLM INT4 (\texttt{cfg1}) lowers the VRAM usage by 21.8–47.5\% but raises the TPOT by 10.4–56.3\% across all platforms (Table~\ref{tab:vram-tpot-final-optimized}). These empirical results demonstrate that within the Jetson Orin Ampere architecture, the dequantization overhead introduced by BitsAndBytes outweighs the native computation speedup provided by INT4 \cite{xu2024survey}. This provides strong evidence supporting H3 for balancing resource savings and inference efficiency. Consequently, LLM INT4 implemented via BitsAndBytes should be viewed purely as a VRAM-saving approach, rather than as a technique for reducing latency.
    \item \textbf{Paradoxical Increase in Energy Consumption.} Despite the decrease in VRAM usage, Qwen3-VL increased energy consumption substantially, i.e., by 54.7\% for NX and 45.3\% for AGX. This increase results from the longer time required for the dequantization operations, which in turn increases power usage. In contrast, LLaVA-OV incurred only a slight increase in energy usage (+1.4\% to +1.9\%). This indicates that the computational cost of dequantization is relatively low for small-scale LLMs (0.49B).
\end{itemize}

\subsection{RQ4: [Interaction Dynamics] Can the accuracy degradation caused by composite quantization configurations (\texttt{cfg3}: Projector INT8 + LLM INT4 and \texttt{cfg5}: Vision INT8 + LLM INT4) be predicted by adding the corresponding single-component ablation results, or is a non-additive pattern observed?}
\label{ssec:res_rq4}
Two configuration pairs are examined to assess additivity:
\begin{itemize}
    \item \small cfg3: $\Delta MME(cfg3) \text{ vs. } \Delta MME(cfg1) \text{ + } \Delta MME(cfg2)$
    \item \small cfg5: $\Delta MME(cfg5) \text{ vs. } \Delta MME(cfg1) \text{ + } \Delta MME(cfg4)$
\end{itemize}

\textcolor{black}{We assess additivity separately for the two composite configurations. For \texttt{cfg3}, the observed degradation is nearly additive across all models, closely matching the linear sum of the individual component errors. In contrast, \texttt{cfg5} deviates from additivity in a model-specific manner. These results lead us to interpret H4 in a more targeted form: composite quantization is near-additive for \texttt{cfg3}, and the hypothesized non-additivity emerges in \texttt{cfg5}, in an architecture-dependent manner.}

\begin{table}[h]
\caption{Additivity analysis: comparison of expected sum of individual quantization impacts ($\Delta$cfg1 + $\Delta$cfg2) with the actual combined impact ($\Delta$cfg3) in NX.}
\label{tab:additivity-analysis}
\vskip -0.05in
\centering
\scriptsize
\setlength{\tabcolsep}{2pt}
\begin{tabularx}{\columnwidth}{@{} l >{\centering\arraybackslash}X >{\centering\arraybackslash}X *{2}{>{\columncolor{gray!15}\centering\arraybackslash}X} >{\centering\arraybackslash}X @{}}
\toprule
\textbf{Model} & \makecell{\textbf{$\Delta$MME}\\\textbf{(cfg1)}} & \makecell{\textbf{$\Delta$MME}\\\textbf{(cfg2)}} & \makecell{\textbf{Expected}\\\textbf{Sum}} & \makecell{\textbf{Actual}\\\textbf{$\Delta$cfg3}} & \makecell{\textbf{Residual}\\\textbf{(cfg3-Exp)}} \\
\midrule
Qwen3-VL & +56.04 & -31.89 & +24.15 & +24.57 & 0.42 \\
DeepSeek-VL2 & +39.50 & +5.86 & +45.36 & +41.80 & -3.56 \\
PaliGemma2 & -62.12 & +2.67 & -59.45 & -60.77 & -1.32 \\
LLaVA-OV & -220.02 & -1.42 & -221.44 & -224.24 & +2.80 \\
Kosmos-2.5 & -10.54 & +1.97 & -8.57 & -8.38 & +0.19 \\
\bottomrule
\end{tabularx}
\end{table}

\begin{table}[h]
\caption{Additivity analysis: comparison of expected sum of individual quantization impacts ($\Delta$cfg1 + $\Delta$cfg4) with the actual combined impact ($\Delta$cfg5) in NX.}
\label{tab:additivity-h5-analysis}
\vskip -0.05in
\centering
\scriptsize 
\setlength{\tabcolsep}{2pt} 
\begin{tabularx}{\columnwidth}{@{} l >{\centering\arraybackslash}X >{\centering\arraybackslash}X *{2}{>{\columncolor{gray!15}\centering\arraybackslash}X} >{\centering\arraybackslash}X @{}}
\toprule
\textbf{Model} & \makecell{\textbf{$\Delta$MME}\\\textbf{(cfg1)}} & \makecell{\textbf{$\Delta$MME}\\\textbf{(cfg4)}} & \makecell{\textbf{Expected}\\\textbf{Sum}} & \makecell{\textbf{Actual}\\\textbf{$\Delta$cfg5}} & \makecell{\textbf{Residual}\\\textbf{(cfg5-Exp)}} \\
\midrule
Qwen3-VL & +56.04 & -48.27 & +7.77 & +7.32 & -0.45 \\
DeepSeek-VL2 & +39.50 & -14.14 & +25.36 & +33.39 & +8.03 \\
PaliGemma2 & -62.12 & +7.27 & -54.85 & -71.18 & -16.33 \\
LLaVA-OV & -220.02 & +7.37 & -212.65 & -212.75 & -0.10 \\
Kosmos-2.5 & -10.54 & +4.18 & -6.36 & -5.63 & +0.73 \\
\bottomrule
\end{tabularx}
\end{table}

\textbf{Key Findings:}
\begin{itemize}
    \item \textbf{Overall Near-Additivity of \texttt{cfg3} (Projector + LLM).} For all models, the \texttt{cfg3} residuals fall within a $\pm 4$-point range, demonstrating nearly additive behavior, as shown in Table~\ref{tab:additivity-analysis}. This indicates minimal interference between the Projector INT8 and LLM INT4 components and confirms that their independent effects overlap linearly.
    \item \textbf{Model-Specific Non-Additivity of \texttt{cfg5} (Vision + LLM).} The behavior of \texttt{cfg5} is non-uniform across models, showing distinct patterns for different architectures. In PaliGemma2, an additional accuracy degradation of 16.33 (NX) is observed in the residuals compared to the expected sum, whereas DeepSeek-VL2 exhibits a super-additive pattern (-8.03), as shown in Table~\ref{tab:additivity-h5-analysis}. These results suggest that the modality-alignment pathways between the vision encoder and the LLM differ across architectures, underscoring the need for prior architecture-specific evaluation for \texttt{cfg5}.
\end{itemize}

\subsection{RQ5: [Hardware Transferability] Is the ordering of the model accuracy across different quantization settings the same on both the Jetson Orin NX and AGX platforms? Are the latency and energy characteristics dependent on the specific platform?}
\label{ssec:res_rq5}
\textcolor{black}{The empirical results demonstrated that the accuracy rankings are strictly preserved across heterogeneous edge platforms regardless of the quantization configuration, thereby obviating the need for hardware-specific accuracy re-validation during deployment (Table~\ref{tab:model-ranking})}.

\begin{table}[h]
\caption{Ranking of evaluated models based on MME total scores across Jetson Orin NX and AGX platforms.}
\label{tab:model-ranking}
\vskip -0.05in
\centering
\scriptsize
\setlength{\tabcolsep}{4pt}
\begin{tabularx}{\columnwidth}{@{} c X X @{}}
\toprule
\textbf{Rank} & \textbf{Jetson Orin NX (Score)} & \textbf{Jetson Orin AGX (Score)} \\
\midrule
1 & Qwen3-VL-2B (2021.78) & Qwen3-VL-2B (1983.32) \\
\addlinespace[0.3em]
2 & DeepSeek-VL2 (1911.12) & DeepSeek-VL2 (1916.71) \\
\addlinespace[0.3em]
3 & PaliGemma2 (1661.01) & PaliGemma2 (1658.10) \\
\addlinespace[0.3em]
4 & LLaVA-OV-0.5B (1361.83) & LLaVA-OV-0.5B (1369.73) \\
\addlinespace[0.3em]
5 & Kosmos-2.5 (662.30) & Kosmos-2.5 (662.05) \\
\bottomrule
\end{tabularx}
\end{table}

In contrast, energy optimization requires a granular understanding of how architectural bottlenecks interact with platform-specific bandwidth constraints. To quantify this, we adopt the IPJ metric, calculated as the normalized MME accuracy per unit of energy consumed (in Joules) \cite{saad2025intelligence}:
$$IPJ(m, h) = \frac{\mathbb{E}_q [acc(m, q)]}{\mathbb{E}_q [P(m, h, q) \cdot \tau(m, h, q)]}$$
where $acc(m, q)$ denotes the normalized MME score ($Score / 2800$) \textcolor{black}{for a given model $m$ and processed query $q$. The denominator represents the energy consumption; specifically, the average power $P(m, h, q)$ is empirically measured using NVIDIA's \texttt{tegrastats} utility at 50ms sampling intervals, and the total inference latency $\tau(m, h, q)$ spans the end-to-end model execution, including vision encoding, projector mapping, the prefill phase, and sequential token generation.}

\begin{table}[h]
\caption{Comparison of intelligence-per-joule (IPJ) using \texttt{cfg0} and \texttt{cfg1} on Jetson Orin platforms.}
\label{tab:ipj-performance}
\vskip -0.05in
\centering
\scriptsize
\setlength{\tabcolsep}{4pt}
\begin{tabularx}{\columnwidth}{@{} l *{4}{>{\centering\arraybackslash}X} @{}}
\toprule
\multirow{2}{*}{\textbf{Models}} & \multicolumn{2}{c}{\textbf{cfg0}} & \multicolumn{2}{c}{\textbf{cfg1}} \\
\cmidrule(lr){2-3} \cmidrule(lr){4-5}
& \textbf{NX} & \textbf{AGX} & \textbf{NX} & \textbf{AGX} \\
\midrule
Qwen3-VL & 5.87 & \textbf{14.45} & 3.90 & \textbf{10.38} \\
\addlinespace[0.2em]
DeepSeek-VL2 & 1.37 & \textbf{3.03} & 0.76 & \textbf{1.72} \\
\addlinespace[0.2em]
PaliGemma2 & 1.65 & \textbf{4.45} & 1.25 & \textbf{3.54} \\
\addlinespace[0.2em]
LLaVA-OV & \textbf{2.27} & 2.25 & 1.86 & \textbf{1.98} \\
\addlinespace[0.2em]
Kosmos-2.5 & 0.12 & \textbf{0.37} & 0.11 & \textbf{0.36} \\
\bottomrule
\multicolumn{5}{l}{\textit{Note: IPJ values represent accuracy per Joule \cite{saad2025intelligence}.}}
\end{tabularx}
\vskip 0.05in
\end{table}

\needspace{4\baselineskip}
\textbf{Key Findings:}
\begin{itemize}
    \item \textbf{Platform-Invariant Accuracy Ranking.} The experimental results confirmed that the MME accuracy ranking by model was completely consistent across all quantization configurations for the two platforms, Jetson NX and AGX, in the order: Qwen3-VL $>$ DeepSeek-VL2 $>$ PaliGemma2 $>$ LLaVA-OV $>$ Kosmos-2.5. These results strongly support hypothesis H5, demonstrating that the architecture of the algorithm has a greater effect on the sensitivity of the models to the quantization settings than the available hardware resources.
    \item \textbf{Asymmetric Power Scaling of Power Consumption Profiles.} Even for the same quantization configuration, NX (TDP 25W) and AGX (TDP 50W) exhibit different power consumption patterns. As shown in Table~\ref{tab:model-ranking}, AGX consistently has a lower average power consumption than NX (ratio 0.65–0.70) on Qwen3-VL. However, on PaliGemma2, LLaVA-OV, and Kosmos-2.5, AGX has a higher average power consumption than NX (ratio 1.02–1.24). These results suggest that the higher TDP (50 W) of AGX does not always lead to higher energy consumption. For Qwen3-VL, the memory bandwidth of AGX is twice as wide (204.8 GB/s), which alleviates computational bottlenecks and results in a lower average power consumption.
    \item \textbf{Platform Specificity and Deployment Strategy for Energy Efficiency (IPJ).} In contrast to the accuracy of task execution, the energy efficiency is highly sensitive to specific model-platform combinations. Aligning with earlier work highlighting the importance of hardware advances \cite{saad2025intelligence}, the present results indicate that Qwen3-VL delivers up to $2.5\times$ higher IPJ on the AGX platform than on the NX, with some models surpassing a $3\times$ gain. Together, these results reinforce the concept that achieving peak efficiency for edge-AI requires an integrated HW-SW co-design strategy, rather than relying solely on standalone model compression.
\end{itemize}

\section{Conclusion}
This study provides important insights into the interaction between the algorithm architecture and hardware deployment by systematically analyzing vision-language models with fewer than 3 billion parameters on heterogeneous edge platforms. The technical implications of these results are as follows:
\textbf{First,} the architecture has a more profound impact on the quantization stability than the model size. Hypothesis H1 predicted severe performance degradation in smaller models. However, consistent performance improvements were observed with MoE-based architectures such as Qwen3-VL. These improvements include an increase of 56.04 in the MME on the NX platform. This suggests that sparse activation functions are effective in localizing and mitigating the propagation of INT4 quantization noise. This provides a clear advantage over dense backbones, such as LLaVA-OV, which exhibited a significant performance decline of 220.02 points.
\textbf{Second,} we confirmed significant hardware-software stack friction in SigLIP-based encoders using BitsAndBytes. Although the accuracy remains consistent, the vision encoding latency increases sharply (by a factor of up to 4.66), suggesting that standard quantization kernels (e.g., BitsAndBytes) are not optimized for the specific computational patterns of SigLIP on Ampere-based edge SoCs. This underscores the need to transition from standalone model compression approaches to hardware-software co-design strategies.

\textbf{Limitations and Future Research.} This study presents a modular framework for PTQ sensitivity analysis, but it is limited to the BitsAndBytes framework. Future research should expand the analysis to include new hardware-native quantization formats, such as FP8, and should evaluate the impact of weight-activation quantization (W8A8) to further reduce the dequantization overhead. Furthermore, exploring automatic precision allocation through neural architecture search (NAS) could enhance the modality-specific deployment guidelines presented in this study.


\section*{Software}
In support of open science and reproducibility, we have released an end-to-end profiling tool. This tool includes a modular quantization framework. The code and related metadata are available in the following repository: \url{https://anonymous.4open.science/r/EdgeQuant-VLMEvalKit/README.md}

\section*{Acknowledgements}
This work was supported by ETRI grant funded by Korean government. [26ZK1100, Honam region strategic industries(optical convergence, energy, AI, etc.) technology advancement and commercialization]


\bibliography{main_ref}
\bibliographystyle{icml2026}

\newpage
\appendix
\onecolumn
\section{Experimental Setup}

\label{app:decoding-hparams}
To support reproducibility, we report the generation hyperparameters used for each VLM wrapper. Model weights are not fine-tuned, and all evaluations share identical VLMEvalKit benchmark prompts, prompt templates, and the same exact-matching answer-extraction protocol, applied uniformly across every model and every quantization configuration. All five models are evaluated with greedy decoding (\texttt{do\_sample}=\texttt{False}); the decoding configuration is held fixed across all quantization settings, so within-model comparisons ($\Delta$MME) are not confounded by decoding differences.

\begin{table}[h]
\centering
\caption{Generation configuration of each VLM wrapper. All models use greedy decoding (\texttt{do\_sample}=\texttt{False}); sampling parameters (temperature, top-$p$, top-$k$) are inapplicable and omitted. Settings are held fixed across all quantization configurations. $^{\dagger}$Kosmos-2.5-chat additionally applies \texttt{no\_repeat\_ngram\_size}$=$\texttt{3} from its released configuration; given MME's short Yes/No answers, this has a negligible effect on accuracy.}
\label{tab:hyperparams}
\vskip 0.1in
\begin{tabular}{lcc}
\toprule
\textbf{Model} & \textbf{Decoding} & \textbf{Max New Tokens} \\
\midrule
Qwen3-VL-2B-Instruct        & Greedy & 16384  \\
DeepSeek-VL2-Tiny           & Greedy & 2048 \\
PaliGemma-3B-mix-448        & Greedy & 512  \\
LLaVA-OneVision-0.5B        & Greedy & 2048 \\
Kosmos-2.5-chat$^{\dagger}$      & Greedy & 1024 \\
\bottomrule
\end{tabular}
\end{table}

\section{Comparison with OpenVLM Leaderboard}

\textcolor{black}{To verify our evaluation pipeline, we compare the MME Total Score produced by our measurement tool (in baseline mode) with the publicly reported scores on the Open VLM Leaderboard \cite{open_vlm_leaderboard}. Table~\ref{tab:leaderboard_comparison} demonstrates that, for the three models (DeepSeek-VL2-Tiny, PaliGemma-3B-mix-448, and LLaVA-OneVision-0.5B), our results closely match the leaderboard scores, supporting the robustness of our evaluation setup. For the remaining two models, the leaderboard and our experiments use different model versions: the leaderboard reports \textbf{Qwen2-VL-2B} and \textbf{Kosmos-2}, whereas our experiments use \textbf{Qwen3-VL-2B-Instruct} and \textbf{Kosmos-2.5-chat} respectively. The score differences for these two models therefore reflect the use of different model versions rather than discrepancies in our evaluation pipeline.}

\begin{table}[h]
\centering
\caption{\textcolor{black}{Comparison of MME Total Score between the Open VLM Leaderboard and our evaluation tool (baseline mode) on two hardware platforms (AGX Orin and Jetson NX). $^\dagger$The leaderboard reports Qwen2-VL-2B, while our experiments use Qwen3-VL-2B-Instruct. $^\ddagger$The leaderboard reports Kosmos-2, while our experiments use Kosmos-2.5-chat.}}
\label{tab:leaderboard_comparison}
\vskip 0.05in
\footnotesize 
\setlength{\tabcolsep}{2pt}
\begin{tabularx}{0.85\columnwidth}{@{} l *{3}{>{\centering\arraybackslash}X} @{}}
\toprule
\textbf{Model} & \makecell[c]{\textbf{Leaderboard}\\\textbf{MME Score}} & \makecell[c]{\textbf{Ours (AGX)}\\\textbf{Baseline}} & \makecell[c]{\textbf{Ours (NX)}\\\textbf{Baseline}} \\
\midrule
Qwen2/3-VL-2B$^\dagger$  & 1899.1 & 1983.32 & 2021.78 \\
DeepSeek-VL2-Tiny        & 1905.5 & 1916.71 & 1911.12 \\
PaliGemma-3B-mix-448     & 1686.1 & 1658.10 & 1661.01 \\
LLaVA-OneVision-0.5B     & 1392.9 & 1359.98 & 1361.83 \\
Kosmos-2.5-chat$^\ddagger$    & 721.1  & 662.05  & 662.30  \\
\bottomrule
\end{tabularx}
\end{table}

\section{MME Total Score across Quantization Configurations}
\label{sec:mme-score-leaderboard-appendix}
\begin{figure*}[t]
  \centering
  \includegraphics[width=\textwidth]{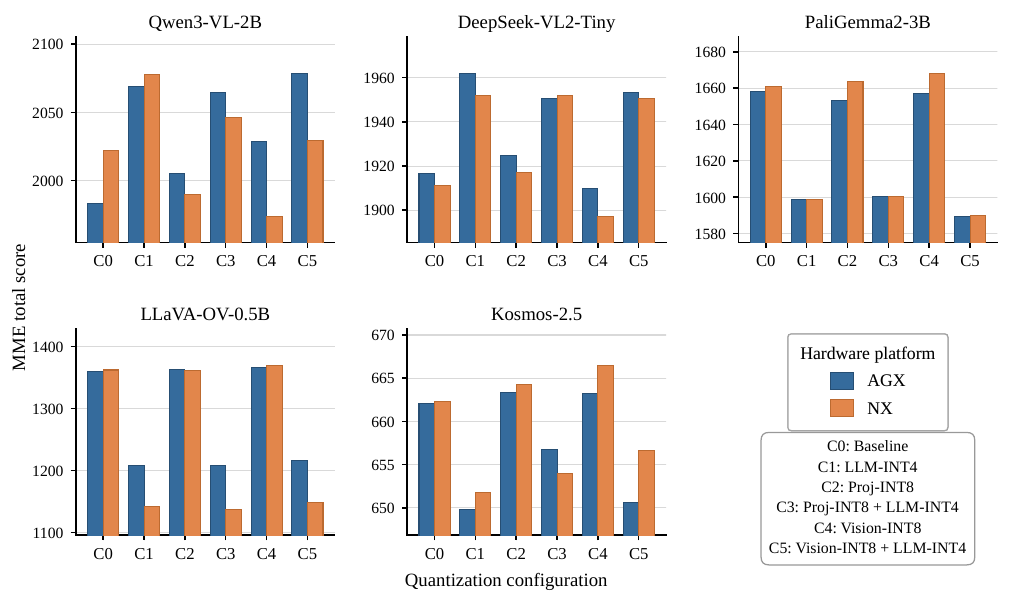}
  \caption{\textcolor{black}{MME total score across quantization configurations (defined in the legend) for five sVLMs on the AGX Orin and Orin NX platforms. Bars show the mean for $n\geq3$ samples; note that each subplot has its own independently scaled y-axis, which does not necessarily start at zero.}}
    \label{fig:mme_total_score}
    \end{figure*}

\textcolor{black}{Figure~\ref{fig:mme_total_score} shows the MME total scores for all five quantization configurations (C0--C5) for each model, as measured on both hardware platforms. Each configuration is evaluated over $n \geq 3$ independent runs, and the outcomes are visualized using bar charts, with each bar representing the mean score across runs. The figure represents the full set of results that are summarized in Table~\ref{tab:final-complete-table}.}

\section{Comprehensive Results across All Quantization Configurations}
\label{app:mme-score-all-appendix}
\textcolor{black}{Table~\ref{tab:vlm_profiling_joint} presents the full set of per-model, per-configuration measurements for all five sVLMs, evaluated under seven quantization configurations (\texttt{cfg0}--\texttt{cfg6}) on both the Jetson Orin NX and Jetson Orin AGX platforms. For each configuration, we report the MME total score, peak VRAM usage~(GB), end-to-end inference latency~(ms), average power draw~(W), and total energy consumption~(J). Configuration definitions follow the notation introduced in Table~\ref{tab:hypotheses-optimized}: \texttt{cfg0} is the FP16 baseline; \texttt{cfg1} applies LLM INT4 only; \texttt{cfg2} applies Projector INT8 only; \texttt{cfg3} combines \texttt{cfg1} and \texttt{cfg2}; \texttt{cfg4} applies Vision INT8 only; \texttt{cfg5} combines \texttt{cfg1} and \texttt{cfg4}; and \texttt{cfg6} applies all three components simultaneously (Vision INT8 + Projector INT8 + LLM INT4). This table constitutes the complete experimental record underlying the summarized results in Table~\ref{tab:final-complete-table}. Across models, LLM INT4 (\texttt{cfg1}) consistently reduces VRAM by approximately 40--50\% but increases latency on both platforms, while Vision INT8 (\texttt{cfg4}) incurs a disproportionate latency penalty for SigLIP-based encoders (PaliGemma-3B-mix-448 and DeepSeek-VL2-Tiny). Energy consumption generally scales with latency, and accuracy degradation under quantization varies substantially across model architectures and configuration types.}
\begin{table*}[t]
\caption{\textcolor{black}{Comprehensive Evaluation of Small VLMs across Quantization Configurations on Jetson Orin NX and AGX.}}
\label{tab:vlm_profiling_joint}
\centering
\scriptsize 
\setlength{\tabcolsep}{1.8pt} 
\begin{threeparttable}
\begin{tabular}{ll ccccc c ccccc}
\toprule
\multirow{2}{*}{\textbf{Model}} & \multirow{2}{*}{\textbf{Config}} & \multicolumn{5}{c}{\textbf{Jetson Orin NX}} & & \multicolumn{5}{c}{\textbf{Jetson Orin AGX}} \\
\cmidrule{3-7} \cmidrule{9-13}
& & \textbf{MME} & \textbf{VRAM} & \textbf{Lat.} & \textbf{Power} & \textbf{Energy} & & \textbf{MME} & \textbf{VRAM} & \textbf{Lat.} & \textbf{Power} & \textbf{Energy} \\
& & \textbf{Score} & \textbf{(GB)} & \textbf{(ms)} & \textbf{(W)} & \textbf{(J)} & & \textbf{Score} & \textbf{(GB)} & \textbf{(ms)} & \textbf{(W)} & \textbf{(J)} \\
\midrule
\multirow{7}{*}{Qwen3-VL-2B-Instruct} 
& cfg0 (Base) & 2021.78 & 4.06 & 908.11  & 13.33 & 12.44 && 1983.32 & 4.06 & 533.85  & 9.07  & 4.91  \\
& cfg1        & 2077.82 & 2.13 & 1406.29 & 13.50 & 19.04 && 2069.05 & 2.13 & 744.09  & 9.76  & 7.08  \\
& cfg2        & 1989.89 & 3.95 & 901.88  & 13.22 & 12.32 && 2005.34 & 3.95 & 537.52  & 8.60  & 4.74  \\
& cfg3        & 2046.35 & 2.04 & 1445.82 & 13.41 & 19.42 && 2064.46 & 2.04 & 801.64  & 9.19  & 7.19  \\
& cfg4        & 1973.51 & 3.77 & 1206.78 & 11.45 & 14.27 && 2028.41 & 3.77 & 738.98  & 7.53  & 5.73  \\
& cfg5        & 2029.10 & 1.85 & 1739.66 & 12.25 & 21.43 && 2078.73 & 1.85 & 984.37  & 8.16  & 8.06  \\
& cfg6 (All)  & 2055.09 & 1.76 & 1761.46 & 12.05 & 21.42 && 2055.09 & 1.75 & 760.82  & 10.65 & 8.06  \\
\midrule
\multirow{7}{*}{DeepSeek-VL2-Tiny} 
& cfg0 (Base) & 1911.12 & 6.82 & 4602.58 & 10.85 & 50.05 && 1916.71 & 6.82 & 2945.87 & 7.63  & 22.57 \\
& cfg1        & 1952.54 & 3.51 & 7883.96 & 11.55 & 91.07 && 1962.10 & 3.51 & 6302.31 & 6.53  & 41.23 \\
& cfg2        & 1916.98 & 6.81 & 4603.13 & 10.87 & 50.14 && 1924.60 & 6.81 & 2958.24 & 7.67  & 22.78 \\
& cfg3        & 1951.55 & 3.51 & 8015.77 & 11.45 & 91.78 && 1950.76 & 3.51 & 6213.22 & 6.57  & 40.91 \\
& cfg4        & 1896.98 & 6.44 & 7482.60 & 9.95  & 74.65 && 1909.74 & 6.44 & 4233.30 & 7.56  & 32.24 \\
& cfg5        & 1953.90 & 3.13 & 10865.29& 10.69 & 116.20&& 1953.11 & 3.13 & 7627.90 & 6.63  & 50.84 \\
& cfg6 (All)  & 1949.64 & 3.12 & 10885.15& 10.67 & 116.26&& 1949.64 & 3.12 & 6098.56 & 7.58  & 46.67 \\
\midrule
\multirow{7}{*}{PaliGemma-3B-mix-448} 
& cfg0 (Base) & 1661.01 & 5.85 & 2440.29 & 14.54 & 35.85 && 1658.10 & 5.85 & 835.30  & 15.53 & 13.28 \\
& cfg1        & 1598.89 & 3.18 & 3127.39 & 14.55 & 45.83 && 1598.74 & 3.18 & 1033.13 & 15.33 & 16.09 \\
& cfg2        & 1663.68 & 5.86 & 2452.48 & 14.53 & 36.01 && 1653.47 & 5.86 & 825.03  & 15.67 & 13.25 \\
& cfg3        & 1600.24 & 3.18 & 3132.63 & 14.55 & 45.88 && 1600.37 & 3.18 & 1053.67 & 15.15 & 16.21 \\
& cfg4        & 1668.28 & 5.49 & 3336.23 & 12.90 & 43.40 && 1656.97 & 5.49 & 1203.12 & 13.16 & 16.09 \\
& cfg5        & 1589.83 & 2.81 & 4027.42 & 13.15 & 53.26 && 1589.19 & 2.81 & 1420.91 & 13.20 & 18.98 \\
& cfg6 (All)  & 1588.56 & 2.81 & 4003.57 & 13.21 & 53.19 && 1588.56 & 2.80 & 1045.41 & 22.38 & 23.78 \\
\midrule
\multirow{7}{*}{LLaVA-OneVision-0.5B} 
& cfg0 (Base) & 1361.83 & 1.89 & 1601.39 & 13.34 & 21.56 && 1369.73 & 2.25 & 1522.36  & 14.25 & 21.74 \\
& cfg1        & 1141.81 & 1.45 & 1701.17 & 13.44 & 23.04 && 1206.15 & 1.76 & 1558.48 & 14.05 & 22.05 \\
& cfg2        & 1360.41 & 1.88 & 1606.90 & 13.33 & 21.63 && 1363.28 & 2.26 & 1531.80 & 14.19 & 21.78 \\
& cfg3        & 1137.59 & 1.45 & 1703.50 & 13.42 & 23.05 && 1207.48 & 1.76 & 1561.79 & 14.14 & 22.24 \\
& cfg4        & 1369.20 & 1.57 & 2751.94 & 10.97 & 30.51 && 1365.30 & 1.91 & 2084.75 & 12.16 & 25.60 \\
& cfg5        & 1149.08 & 1.13 & 2859.89 & 11.09 & 31.99 && 1215.45 & 1.40 & 2130.10 & 12.06 & 26.06 \\
& cfg6 (All)  & 1215.94 & 1.12 & 2883.24 & 11.04 & 32.19 && 1215.94 & 1.40 & 1558.00 & 20.49 & 32.39 \\
\midrule
\multirow{7}{*}{Kosmos-2.5-chat} 
& cfg0 (Base) & 662.30  & 6.45 & 16225.28& 12.11 & 196.82&& 662.05  & 6.45 & 4181.44 & 14.79 & 62.17 \\
& cfg1        & 651.76  & 5.59 & 16495.28& 12.25 & 202.37&& 649.84  & 5.59 & 4300.84 & 14.74 & 63.68 \\
& cfg2        & 664.27  & 6.45 & 16227.37& 12.12 & 196.98&& 663.30  & 6.45 & 4196.07 & 14.90 & 62.84 \\
& cfg3        & 653.92  & 5.58 & 16488.09& 12.28 & 202.72&& 656.76  & 5.58 & 4334.14 & 14.93 & 65.03 \\
& cfg4        & 666.48  & 5.98 & 17773.39& 11.95 & 212.64&& 663.27  & 5.98 & 4748.06 & 14.25 & 67.98 \\
& cfg5        & 656.67  & 5.13 & 18046.58& 12.09 & 218.46&& 650.59  & 5.13 & 4864.09 & 14.10 & 68.89 \\
& cfg6 (All)  & 652.51  & 5.12 & 18152.66& 12.07 & 219.40&& 652.51  & 5.12 & 3721.97 & 24.38 & 91.22 \\
\bottomrule
\end{tabular}
\begin{tablenotes}[flushleft]
    \scriptsize
    \item[] \textcolor{black}{\textbf{Note:} cfg0 (FP16 base), cfg1 (LLM INT4), cfg2 (Proj INT8), cfg3 (cfg1+cfg2), cfg4 (Vis INT8), cfg5 (cfg1+cfg4), and cfg6 (Vis INT8 + Proj INT8 + LLM INT4). Metrics default to: Latency (ms), Avg. Power (W), and Energy (J).}
\end{tablenotes}
\end{threeparttable}
\end{table*}

\FloatBarrier
\FloatBarrier

\section{Isolated Vision-Path Profiling for the SigLIP INT8 Latency Anomaly}
\label{app:vision-int8-control}
To isolate the vision INT8 anomaly from the host LLM, projector, tokenizer, and decoding loop, we profile the vision path alone using NVTX-annotated ranges with NVIDIA Nsight Systems (50 iterations; Nsight Systems 2026.3.1.157). For PaliGemma2 and Kosmos-2.5 we measure the outer \texttt{isolated\_vision\_encoder} range. For DeepSeek-VL2-Tiny, whose standalone vision call does not reproduce its multimodal path, we instead profile the generation-free \texttt{prepare\_inputs\_embeds} path. We note that this difference in measured scope means the DeepSeek-VL2-Tiny numbers are not strictly comparable to the encoder-only ranges of the other two models; we report it separately and interpret it qualitatively rather than in direct head-to-head comparison. As shown in Table~\ref{tab:vision-int8-control}, the large slowdown is reproduced for SigLIP-style encoders on both platforms (PaliGemma2: $4.93\times$ on AGX, $4.32\times$ on NX; DeepSeek-VL2-Tiny: $3.10\times$), confirming that the anomaly is not an artifact of the decoding loop or a single AGX-specific measurement. The non-SigLIP Kosmos-2.5 control, by contrast, slows by only $1.19\times$ (AGX) and $1.16\times$ (NX).

Nsight kernel summaries explain this contrast (Table~\ref{tab:kernel_breakdown}). Under BitsAndBytes vision INT8, the PaliGemma2 vision path is restructured into repeated cycles of activation quantization, INT8 CUTLASS GEMM, dequantization, and synchronization. Quantization and dequantization kernels alone account for $22.8$--$25.7\%$ of listed INT8 GPU kernel time (NX and AGX, respectively), accompanied by a $7.4\times$ rise in kernel launches and a sharp increase in memory copies. This points to a fragmented execution path rather than a faster GEMM substitution. The non-SigLIP Kosmos-2.5 control, already dominated by softmax, copy, and elementwise kernels, dilutes the same overhead to $6.0$--$6.5\%$. We restrict the kernel-level breakdown in Table~\ref{tab:kernel_breakdown} to these two models, as the latency results for DeepSeek-VL2-Tiny are reported only on AGX.

These results lead us to revise H2 toward a narrower, deployment-specific interpretation: the disproportionate slowdown arises from the interaction between SigLIP-style encoders, the BitsAndBytes INT8 execution path, and the Jetson Orin Ampere architecture, rather than from any inherent inefficiency of SigLIP or of INT8 vision quantization in general.

\begin{table}[t]
\caption{Impact of BitsAndBytes vision INT8 quantization on isolated vision-encoding latency (ms), with a non-SigLIP Kosmos-2.5 control. Ratio $=$ Vision INT8 / FP16.}
\label{tab:vision-int8-control}
\vskip 0.15in
\begin{center}
\begin{small}
\setlength{\tabcolsep}{4pt}
\begin{tabular}{l l l S[table-format=5.2] S[table-format=5.2] c}
\toprule
Model & Plat. & Encoder & {FP16 (ms)} & {INT8 (ms)} & {Ratio $\uparrow$} \\
\midrule
PaliGemma2        & AGX    & SigLIP-So400m & 195.99   & 966.09   & $\times$4.93 \\
PaliGemma2        & NX     & SigLIP-So400m & 313.14   & 1353.53  & $\times$4.32 \\
DeepSeek-VL2-Tiny & AGX    & SigLIP-style  & 1145.81  & 3555.38  & $\times$3.10 \\
Kosmos-2.5          & AGX    & Pix2Struct-style    & 7058.22  & 8373.13  & $\times$1.19 \\
Kosmos-2.5          & NX     & Pix2Struct-style    & 10701.11 & 12428.37 & $\times$1.16 \\
\bottomrule
\end{tabular}
\end{small}
\end{center}
\vskip -0.1in
\end{table}

\begin{table}[t]
\centering
\caption{Kernel-level breakdown of BitsAndBytes vision INT8 on the isolated vision path. Quant/dequant share is the fraction of listed INT8 GPU kernel time spent in \texttt{kInt8VectorQuant} (activation quantization) and \texttt{kdequant\_mm\_int32\_fp16} (dequantization). Kernel-launch ratio is INT8 relative to FP16. DeepSeek-VL2-Tiny is omitted here because its latency is reported on AGX only (Table 14); for reference, its AGX quant/dequant share is 22.5\%.}
\label{tab:kernel_breakdown}
\begin{tabular}{lllrr}
\toprule
Model & Plat. & Encoder & Q/Dq share & Launches $\uparrow$ \\
\midrule
PaliGemma2 & AGX & SigLIP     & 25.7\% & 7.4$\times$ \\
PaliGemma2 & NX  & SigLIP     & 22.8\% & 7.4$\times$ \\
Kosmos-2.5   & AGX & Pix2Struct-style & 6.5\%  & 2.8$\times$ \\
Kosmos-2.5   & NX  & Pix2Struct-style & 6.0\%  & 2.8$\times$ \\
\bottomrule
\end{tabular}
\end{table}

\section{Non-BitsAndBytes AWQ Sanity Check}
\label{app:awq-sanity}
To examine whether the LLM INT4 latency trend reported in RQ3 (Section~\ref{ssec:res_rq3}) is specific to the BitsAndBytes backend, we additionally evaluate a non-BitsAndBytes AWQ baseline on Qwen3-VL-2B-Instruct (Table~\ref{tab:awq-sanity}). This experiment is intended as a sanity check rather than a full replacement for the component-wise BitsAndBytes study. We apply AWQ W4A16 exclusively to the text-decoder Linear layers, excluding the vision encoder, the visual merger, and the lm\_head, with calibration on 512 held-out Flickr30k image-text samples. For the AWQ sanity-check run, we report median TPOT rather than relying only on mean end-to-end latency, because a small number of Qwen3-VL samples produced unusually long responses that would otherwise inflate the mean end-to-end latency.
The AWQ baseline nearly preserves the MME accuracy of the FP16/BF16 baseline and, unlike BitsAndBytes INT4, does not reproduce the TPOT penalty, indicating that the TPOT increase under BitsAndBytes LLM INT4 is tied to the BitsAndBytes execution path in our runtime, rather than being an inherent property of 4-bit LLM quantization. However, AWQ does not reduce the peak inference VRAM in our current runtime. This suggests that peak allocation during multimodal inference is likely dominated by runtime activations, visual features, the KV cache, and the compressed-tensor execution workspace, rather than by the quantized weight storage alone. We therefore report AWQ as a non-BitsAndBytes sanity-check backend, not as a substitute for the component-wise BitsAndBytes analysis that forms the basis of this study.

\begin{table*}[h]
\caption{Non-BitsAndBytes AWQ sanity check on Qwen3-VL-2B-Instruct, compared with the FP16/BF16 baseline and BitsAndBytes (BnB) LLM INT4.}
\label{tab:awq-sanity}
\vskip 0.1in
\begin{center}
\begin{small}
\setlength{\tabcolsep}{4pt}
\begin{tabular}{l l l l r r r r}
\toprule
Model & Backend & Quant. comp. & Calib. & MME & Peak VRAM & Median TPOT \\
\midrule
Qwen3-VL-2B & FP16/BF16  & None & None          & 2021.78 & 4.06~GB & 111.1~ms \\
Qwen3-VL-2B & BnB INT4   & LLM  & None          & 2077.82 & 2.13~GB & 173.1~ms \\
Qwen3-VL-2B & AWQ W4A16  & LLM  & Flickr30k 512 & 2020.93 & 4.08~GB & 107.5~ms \\
\bottomrule
\end{tabular}
\end{small}
\end{center}
\end{table*}
\FloatBarrier


\end{document}